\documentclass[10pt,twocolumn,letterpaper]{article}

\usepackage{iccv}
\usepackage{times}
\usepackage{epsfig}
\usepackage{graphicx}
\usepackage{amsmath}
\usepackage{amssymb}

\usepackage{caption}
\usepackage{xspace}
\usepackage{times}
\usepackage{url}
\usepackage{multirow}
\usepackage{epstopdf}
\usepackage{amsmath}
\usepackage[pagebackref=true,breaklinks=true,letterpaper=true,colorlinks,bookmarks=false]{hyperref}

\iccvfinalcopy % *** Uncomment this line for the final submission

\def\iccvPaperID{202} % *** Enter the ICCV Paper ID here

\ificcvfinal\fi
\begin{document}

%%%%%%%%% TITLE
\title{Resolving Scale Ambiguity Via XSlit Aspect Ratio Analysis}

\author{Wei Yang\\
\and
Haiting Lin\text{$^1$}\\
\and
Sing Bing Kang\text{$^2$}\\
\and
Jingyi Yu\text{$^1$}\\
\and
$^1$University of Delaware\\
{\tt\small\{wyangcs,haiting,jingyiyu\}@udel.edu}
\and
$^2$Microsoft Research\\
{\tt\small sbkang@microsoft.com}
}

%\author{Wei Yang\\
%University of Delaware\\
%Newark, DE, USA, 19716\\
%{\tt\small wyangcs@udel.edu}
%% omit the following lines up until the closing ``}''.
%% Additional authors and addresses can be added with ``\and'',
%% just like the second author.
%% To save space, use either the email address or home page, not both
%\and

\maketitle
%\thispagestyle{empty}

%%%%%%%%% ABSTRACT
\begin{abstract}
    In perspective cameras, images of a frontal-parallel 3D object preserve its aspect ratio invariant to its depth. Such an invariance is useful in photography but is unique to perspective projection. In this paper, we show that alternative non-perspective cameras such as the crossed-slit or XSlit cameras exhibit a different depth-dependent aspect ratio (DDAR) property that can be used to 3D recovery. We first conduct a comprehensive analysis to characterize DDAR, infer object depth from its AR, and model recoverable depth range, sensitivity, and error. We show that repeated shape patterns in real Manhattan World scenes can be used for 3D reconstruction using a single XSlit image. We also extend our analysis to model slopes of lines. Specifically, parallel 3D lines exhibit depth-dependent slopes (DDS) on their images which can also be used to infer their depths. We validate our analyses using real XSlit cameras, XSlit panoramas, and catadioptric mirrors. Experiments show that DDAR and DDS provide important depth cues and enable effective single-image scene reconstruction.
\end{abstract}

%  Ko Nishino
%1. \cite{novatnack2007scale}: present a comprehensive framework for exploiting the 3D geometric scale variability. Specifically, we focus on detecting scale-dependent geometric features on triangular mesh models of arbitrary topology.
%1. \cite{novatnack2008scale}: align range images with the same global scale using scale-dependent local shape descriptors.

%%%%%%%%% BODY TEXT
\section{Introduction}

%Even knowing the relative ratios of the object, we can not conduct scale estimations.

\begin{figure}
\includegraphics[width=0.9\linewidth]{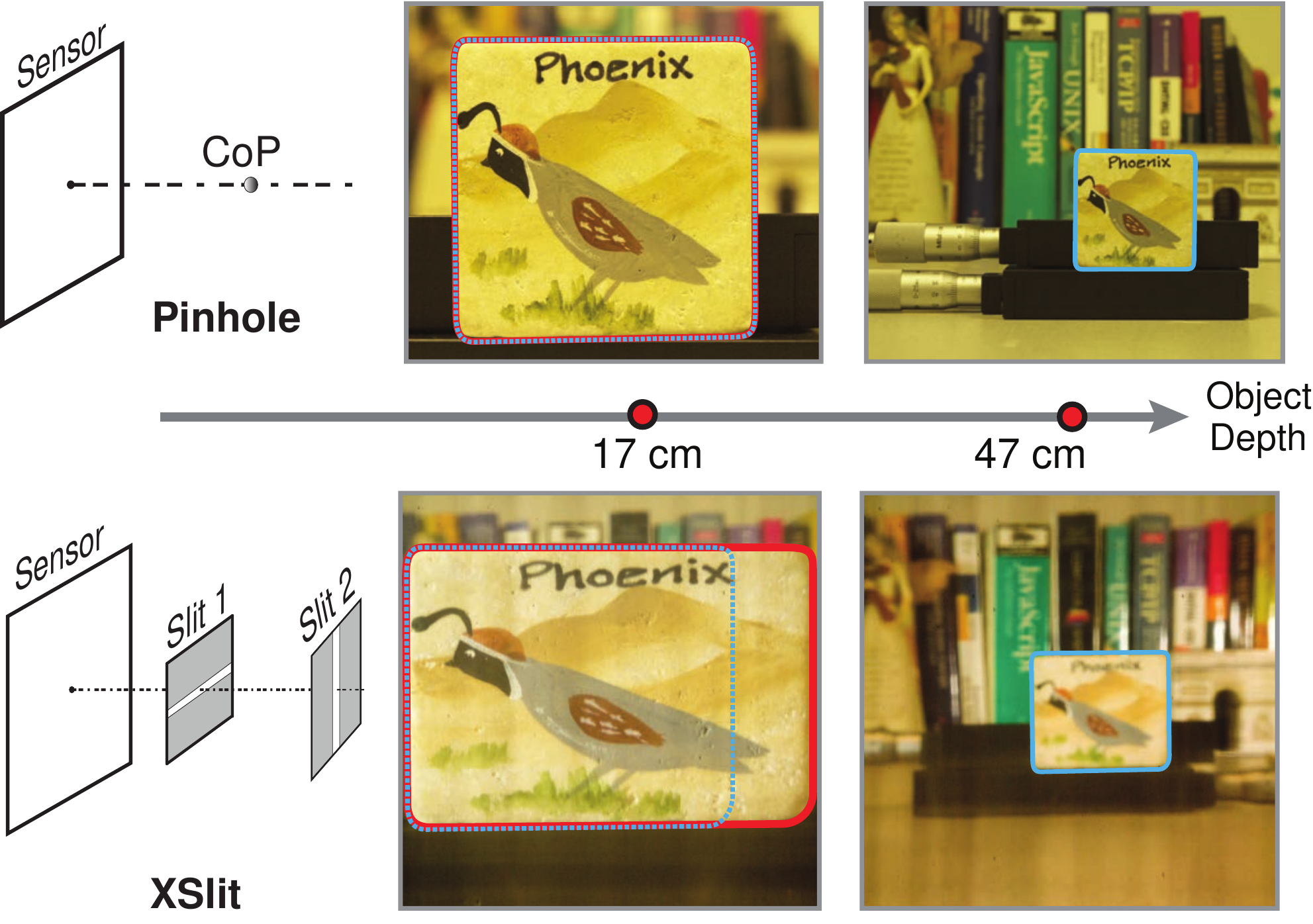}
\centering \caption{Images of the same object lying at different depths have an identical aspect ratio (AR) in a perspective camera (Top) but have very different ARs in an XSlit image (Bottom).}
\label{fig:teaser}
\end{figure}

\begin{figure}
\includegraphics[width=1.0\linewidth]{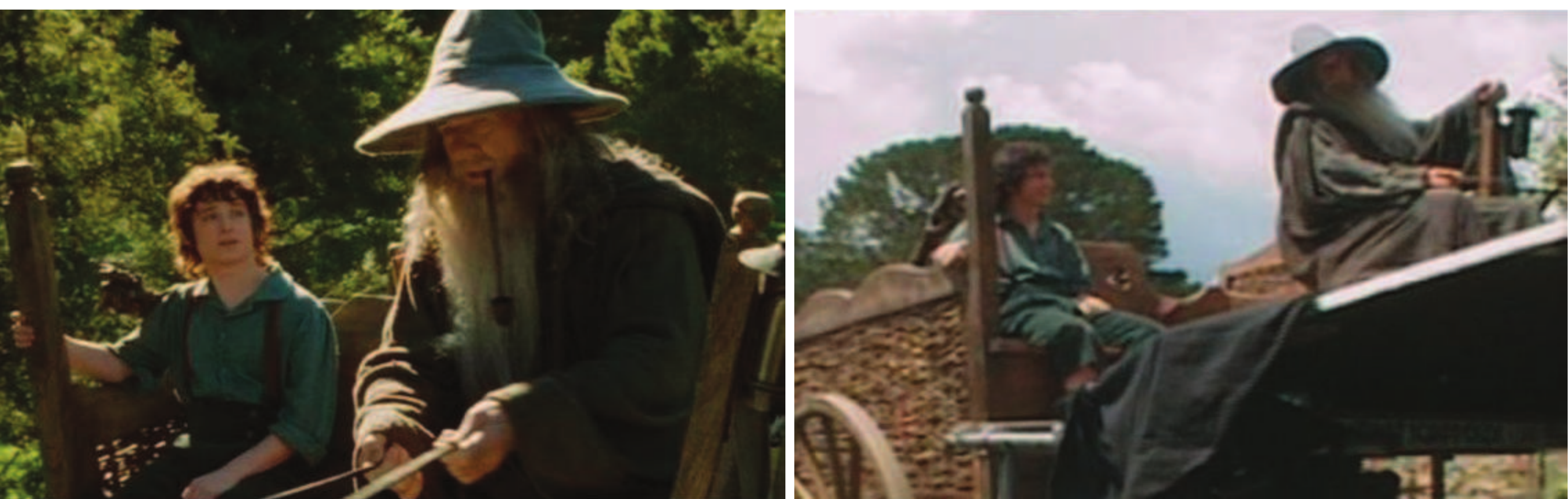}
\centering \caption{The perspective trick used in the movie ``The Lord of the Rings''.}
\label{fig:LoR}
\end{figure}

A single perspective image exhibits scale ambiguity: 3D objects of difference sizes can have images of an identical size under perspective projection, as shown in Fig.~\ref{fig:teaser}. In photography and architecture, the forced perspective technique employs this optical illusion to make an object appear farther away, closer, larger or smaller than its actual size while preserving the aspect ratio. Fig.~\ref{fig:LoR} shows an example in the film ``the Lord of the Rings'' where characters apparently standing next to each other would be displaced by several feet in depth from the camera. For computer vision, however, such an invariance provides little help, if not harm, to scene reconstruction.

Prior approaches on resolving the scale ambiguity range from imposing shape priors \cite{cabral2014piecewise, furukawa2009reconstructing}, extracting local descriptors \cite{novatnack2007scale} to analyzing the vanishing points \cite{kovsecka2002video}. In this paper, we approach the problem from a different angle: we analyze aspect ratio changes of an object with respect to its depth. Consider a frontal-parallel rectangle $R$ of size $l_h \times l_v$ located $d$ away from the sensor and $d > f$ where $f$ is the camera's focal length. Under perspective projection, its image is an rectangle $R^\prime$ similar to R of size $[l'_h, l'_v]=\frac{f}{d-f}[l_h, l_v]$. This implies that the aspect ratio $r=l_v/l_h$ of $R$ and $R'$ remain the same. The property can termed as aspect-ratio invariance (ARI). ARI is an important property of perspective projection. ARI, however, no longer holds under non-centric projections, exhibiting depth-dependent aspect-ratio (DDAR).

In this paper, we explore DDAR in a special type of non-centric cameras called the crossed-slit or XSlit camera \cite{Zomet09XSlit}. Earlier work in XSlit imaging includes the pushbroom camera used in satellite imaging and XSlit panoramas by stitching a sequence of perspective images. The General Linear Camera theory \cite{Yu2004GLC} has shown that the XSlit camera is generic enough to describe a broad range of non-centric cameras. In fact, pushbroom, orthographic and perspective cameras can all be viewed as special XSlit entities. Geometrically, an XSlit camera collects rays that simultaneously pass through two oblique (neither parallel nor coplanar) slits in 3D space, in contrast to a pinhole camera whose rays pass through a common 3D point. Ye et al.\cite{Ye14Rotation} has further proposed a practical realization by relaying a pair of cylindrical lenses coupled with slit-shaped apertures.

We show that the XSlit camera exhibits DDAR that can help resolve scale ambiguity. Consider two 3D rectangles of an identical size lying at different depth with their images being $R_1$ and $R_2$ respectively. Different from the pinhole case, the AR of $R_1$ and $R_2$ will be different, as shown in Fig.~\ref{fig:teaser}. We first develop a comprehensive analysis to characterize DDAR in the XSlit camera. This derivation leads to a simple but effective graph-cut based scheme to recover object depths from a single XSlit image and an effective formulation to model recoverable depth range, sensitivity, and errors. In particular, we show how to exploit repeated shape patterns exhibiting in real Manhattan World scenes to conduct 3D reconstruction.

Our DDAR analysis can further be extended to model the slopes of lines. Specifically, for parallel 3D lines of a common direction, we show that as far as the direction is different from both slits, their projections will exhibit depth-dependent slopes or DDS, i.e., the projected 2D lines will have different slopes depending on their depths.  DDS and DDAR can be combined to further improve 3D reconstruction accuracy. We validate our theories and algorithms on both synthetic and real data. For real scenes, we experiment on different types of XSlit images including the ones captured by the XSlit lens \cite{Ye14Rotation} and synthesized as stitched panoramas \cite{schechner2001generalized}. In addition, our scheme can be applied to catadioptric mirrors by modeling reflections off the mirrors as XSlit images. Experiments show that DDAR and DDS provide important depth cues and enable effective single-image scene reconstruction.

\section{Related Work}

Our work is most related to Manhattan World reconstruction and non-centric imaging.

A major task of computer vision is to infer 3D geometry of scenes using as fewer images as possible. Tremendous efforts have focused on recovering a special class of scene called the Manhattan World (MW) \cite{coughlan1999manhattan}. MW is composed of repeated planar surfaces and parallel lines aligned with three mutually orthogonal principal axes and fits well to many man-made (interior/exterior) environments. Under the MW assumption, one can simultaneously conduct 3D scene reconstruction \cite{delage2006dynamic,furukawa2009reconstructing} and camera calibration \cite{schindler2004atlanta}.

MW generally exhibits repeated line patterns but lacks textures and therefore traditional stereo matching is less suitable for reconstruction. Instead, prior-based modeling is more widely adopted. For example, Furukawa \etal \cite{furukawa2009reconstructing} assign a plane to each pixel and then apply graph-cut on discretized plane parameters. Other monocular cues such as the vanishing points \cite{criminisi2000single} and the reference planes (\eg the ground) have also been used to better approximate scene geometry. Hoime \etal \cite{hoiem2005geometric,hoiem2005automatic} use image attributes (color, edge orientation, \etc) to label image regions with different geometric classes (sky, ground, and vertical) and then ``pop-up" the vertical regions to generate visually pleasing 3D reconstructions. Similar approaches have been used to handle indoor scenes \cite{delage2006dynamic}. Machine learning techniques have also been used to infer depths from image features and the location and orientation of planar regions \cite{saxena2005learning,saxena2007learning}. Lee \etal \cite{lee2009geometric} and Flint \etal \cite{flint2010dynamic} search for the most feasible combination of line segments for indoor MW understanding.

Our paper explores a different and previously overlooked properties of MW: the scene contains multiple objects with an identical aspect ratio or size (e.g., windows) but lie at different depths. In a perspective view, these patterns will map to 2D images of an identical aspect ratio. In contrast, we show that the aspect ratio changes with respect to depth if one adopts a non-centric or multi-perspective camera. Such imaging models widely exist in nature, e.g., a compound insect eye, reflections and refractions of curved specular surfaces, images seen through volumetric gas such as a mirage, etc. Rays in these cameras generally do not pass through a common CoP and hence do not follow pinhole geometry. Consequently, they lose some nice properties of the perspective camera (e.g., lines no longer project to lines); at the same time they also gain some unique properties such as the coplanar common points \cite{yang2014coplanar}, special shaped curves \cite{ye2013manhattan}, etc. In this paper, we focus on the depth-dependent aspect ratio (DDAR) property for inferring 3D geometry.

The special non-centric camera we employ here is the crossed-slit or XSlit camera. An XSlit camera collects rays simultaneously passing through two oblique lines (slits) in 3D space. The projection geometry of an XSlit has been examined in various forms in previous studies, \eg, as projection model in \cite{Zomet09XSlit}, as general linear constraints in \cite{Yu2004GLC}, and as ray regulus in \cite{ponce2009camera}. For long the XSlit camera has been restricted to a theoretical model as it is physically difficult to acquire ray geometry following the slit structure. The only exception is the XSlit panoramas \cite{seitz2002space,pajdla2002geometry} where an XSlit panorama can be stitched from a translational sequence of images or more precisely a 3D light field \cite{Levoy1996LFR}. Recently, Ye et al.\cite{Ye14Rotation} presented a practical XSlit camera. Their approach relays two cylindrical lenses with perpendicular axes, each coupled with a slit shaped aperture to achieve in-focus imaging.

%The problem is received as a single-image depth recovery problem. Existing solutions can be categorized into 1) prior based and 2) (camera) geometry based. Prior based methods include shape from shading/texture [XXX], or more sophisticated learned feature cues[XXX]. Camera geometry based methods include scene structure from vanishing points [XXX], and intersection points~\cite{ramalingam2013lifting}. However, these methods use pinhole camera and all suffer from scale ambiguity: only a relative geometry is recoverable. Recently, Ye {\it et al.}~\cite{ye2013a} propose to use a multi-perspective camera (XSlit camera) to conduct Manhattan scene geometry understanding. They explore the unique property of Coplanar Common Point (CCP) and combine with XSlit Vanishing Point (XVP) to detect planes of a Manhattan scene.
%------------------------------------------------------------------------
\section{Depth Dependent Aspect Ratio}
We first analyze how aspect ratio of an object changes with respect to its depth in an XSlit camera. We call this property Depth-Dependent Aspect Ratio or DDAR.

%------------------------------------------------------------------------
\subsection{XSlit Camera Geometry}

\begin{figure}
\includegraphics[width=0.9\linewidth]{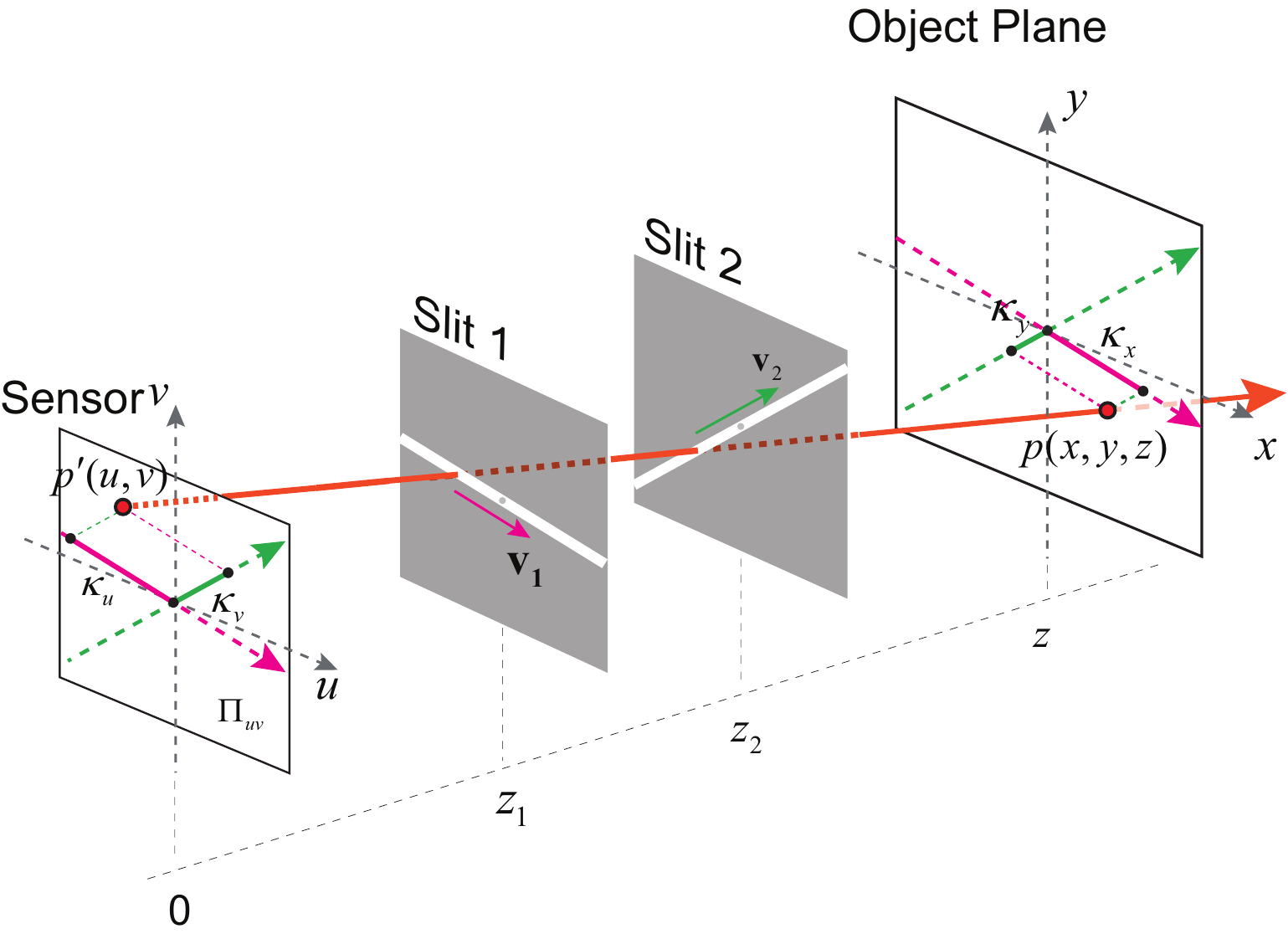}
\centering \caption{ XSlit camera geometry: rays collected by the camera should simultaneously pass through two slits at different depths.}
\label{fig:xslitgeo}
\end{figure}

A XSlit camera collects rays that pass through two oblique slits (neither coplanar nor parallel) simultaneously. For simplicity, we align the sensor plane to be parallel to both slits and corresponds to the x-y plane. Such a setup is consistent with the real XSlit design \cite{Ye14Rotation} and the XSlit panoramas \cite{Zomet09XSlit}. Further, we assume the origin of the coordinate system corresponds to the intersection of the two slits\text{'} orthogonal projections on the sensor plane, as shown in Fig.~\ref{fig:xslitgeo}. The two slits lie at depth $z_1$ and $z_2$ and have angle $\theta_1$ and $\theta_2$ w.r.t the $x$-axis, where $z_2 > z_1$ and $\theta_1 \neq \theta_2$. Under this setup, the $z$ components along the two slits are 0. And the $x$-$y$ directions are $\mathbf{v_1}[\cos \theta_1, \sin \theta_1]$ and $\mathbf{v_2}[\cos \theta_2, \sin \theta_2]$ that spans $\mathbf{R}^2$ space.

Previous approaches study projection using XSlit projection matrix \cite{Zomet09XSlit}, light field parametrization\cite{Yu2004GLC}, and linear oblique\cite{pajdla2002geometry}. Since our analysis focuses on aspect ratio, we introduce a simpler projection model analogous to pinhole projection. Consider a 3D point $p$ to $p^\prime$. The process can be described as follows: first decompose the $x$-$y$ components of $p$ into two basis vectors, $\mathbf{v_1}$, $\mathbf{v_2}$ and write it as $[\kappa_x, \kappa_y, z]$. Next project individual component to $[\kappa_u, \kappa_v]$. Each component can be viewed as pinhole projection as they are parallel to either slits. Finally obtain the mapping from $p$ to $p^\prime$.

We first represent $p$ on the basis of $\mathbf{v_1}$ and $\mathbf{v_2}$

%Notice that the $v_1$ and $v_2$ can also be viewed as basis vectors in $\mathbf{R}^2$ space. Therefore we can treat them as the virtual $x$ and $y$ axes and represent every 3D point $\mathbf{p}(x,y,z) = XXX$:

\begin{equation*}
\label{eq:kxky}
\left [
\begin{matrix}
x \\
y
\end{matrix}
\right ]
=
\kappa_x \mathbf{v_1}
+
\kappa_y \mathbf{v_2}
\end{equation*}
%where $\kappa$ XXX means what.

%Recall that the XSlit projection $\Phi$ satisfies $\Phi(a+b) = \Phi(a)+\Phi(b)$. Hence
We then project $\kappa_x \mathbf{v_1}$ and $\kappa_y \mathbf{v_2}$ independently. Notice the two components are at depth $z$. And $\kappa_x \mathbf{v_1}$ is parallel to slit 1 and $\kappa_y \mathbf{v_2}$ is parallel to slit 2. Their projections imitate the pinhole projection except that the focal lengths are different:

\begin{equation}
\label{eq:pinproj}
\kappa_u = -\frac{z_2}{z-z_2}\kappa_x, \kappa_v = -\frac{z_1}{z-z_1}\kappa_y
\end{equation}

Notice the XSlit mapping is linear, we can combine $\kappa_u$ and $\kappa_u$ to compute $p^\prime$.

\begin{equation*}
\label{eq:pprim}
p^\prime = \kappa_u \mathbf{v_1} + \kappa_v \mathbf{v_2}
\end{equation*}

$\kappa_u$ and $\kappa_v$ are also the linear representations of $p^\prime$ on basis of $\mathbf{v_1}$ and $\mathbf{v_2}$.

%Divide the two equations, we have:
%\begin{equation}
%\label{eq:glcccpeqn}
%\frac{\kappa_u}{\kappa_v} = \frac{z_2(z-z_1)}{z_1(z-z_2)}\frac{\kappa_x}{\kappa_y}
%\end{equation}
%
%Here we define $r_0 = \frac{\kappa_x}{\kappa_y}$ is the aspect ratio of the original object. And $r_i = \frac{\kappa_u}{\kappa_v}$ is the aspect ratio in the image.

%\textbf{Linear space:}

%------------------------------------------------------------------------
\subsection{Aspect Ratio Analysis}

Equation ~\ref{eq:pinproj} reveals that $\kappa_x$ and $\kappa_y$ are projected to $\kappa_u$ and $\kappa_v$ with different scale on the two directions parallel to the slits. In other words, with the change of depth, the ratio will be change accordingly. Specifically, we can compute the ratio as:

\begin{equation}
\label{eq:ardef}
\frac{\kappa_u}{\kappa_v} = \frac{z_2(z-z_1)}{z_1(z-z_2)}\frac{\kappa_x}{\kappa_y}
\end{equation}

This is fundamentally different from the pinhole/perspective case where the ratio remains static across depth. To understand why it is the case, recall that the pinhole camera can be viewed as a special XSlit camera where the two slits intersect, i.e., they are at the same depth $z_1 = z_2$. In that case, Eqn. degenerates to $\frac{\kappa_x}{\kappa_y} = \frac{\kappa_u}{\kappa_v}$, i.e., the aspect ratio is invariant to depth.

For the rest of the paper, we use $r_o = \frac{\kappa_x}{\kappa_y}$ to represent the base aspect ratio and $r_i = \frac{\kappa_u}{\kappa_v}$ represents the aspect ratio after XSlit projection. From Eqn. ~\ref{eq:ardef}, we can derive the depth from the aspect ratio as:

\begin{equation}
\label{eq:getz}
z=\frac{z_1z_2(r_i-r_o)}{z_1r_i-z_2r_o}
\end{equation}

\paragraph{Monotonicity:} Given a fixed XSlit camera, Eqn. ~\ref{eq:getz} reveals that the AR monotonically decreases with respect to $z$. In fact, we can compute the derivative of $z$ with respect to $r_i$:
\begin{equation}
\label{eq:glcccpeqn}
\frac{\partial{z}}{\partial{r_i}} = \frac{z_1z_2(z_1-z_2)r_o}{(z_1r_i-z_2r_o)^2}
\end{equation}
Since $z_1<z_2$, we have $\frac{\partial{z}}{\partial{r_i}} < 0$, i.e., the depth $z$ decrease monotonically with $r_i$. In fact the minimum and the maximum ARs correspond to:

\begin{equation}
\label{eq:ratiorange}
r_i^{\min} = r_i|_{z \rightarrow \infty} = \frac{z_2}{z_1}r_o, r_i^{\max}= r_i|_{z \rightarrow z_2} = \infty
\end{equation}

\paragraph{Depth Sensitivity:} Another important we address here is depth sensitivity. We compute the partial derivative of $r_i$ respect to $z$ for $z$ ranging from $z_2$ to $\infty$ and we have:

\begin{equation}
\label{eq:sensit}
\frac{\partial{r_i}}{\partial{z}} = \frac{z_2(z_1-z_2)}{z_1(z-z_2)^2}r_o
\end{equation}

The sensitivity is the absolute value of $\frac{\partial{r_i}}{\partial{z}}$ and it decrease monotonically for $z>z_2$. This implies that as objects get further away, the depth accuracy recoverable from the AR also decreases. According to Eqn. ~\ref{eq:sensit}, the sensitivity is positively related to $\frac{z_2}{z_1}$ and $z_1 - z_2$. Farther separated slits and greater ratio between two slits distances corresponds to higher sensitivity. This phenomenon resembles classical stereo matching using two perspective cameras where the deeper the object, the smaller the disparity and the less accuracy that stereo matching can produce.

\paragraph{Depth Range:} We can further compute the maximum discernable depth $z^{\max}$. To do so, we first compute $r_i$ when $z\rightarrow\infty$ as $r_i^\infty = \frac{z_2}{z_1}r_o$. Next we change $r_i^\infty$ with $\epsilon$, the smallest ratio change that is discernable in image. We have $r_i^\ast=\frac{z_2}{z_1}r_o + \epsilon$. The lower bound of $\epsilon$ is $1/L$, $L$ is the image width or height, without considering subpixel accuracy. Sine the depth changes monotonically with $r_i$, the maximum discernable depth is correspond to $r_i^\ast$. Finally we compute the depth use Eqn. ~\ref{eq:getz}:

\begin{equation}
\label{eq:depthrange}
z^{\max} = \frac{z_2}{z_1}[ 1+ (z_2 - z_1)\frac{r_o}{\epsilon} ]
\end{equation}
Eqn. ~\ref{eq:depthrange} indicates that the larger slit distance ratio $\frac{z_2}{z_1}$ and bigger separating distance of two slits $z_2 - z_1$ correspond to a larger discernable depth range.

\section{Depth Inference from DDAR}
\label{sec:depthinfer}

\begin{figure}
\includegraphics[width=0.9\linewidth]{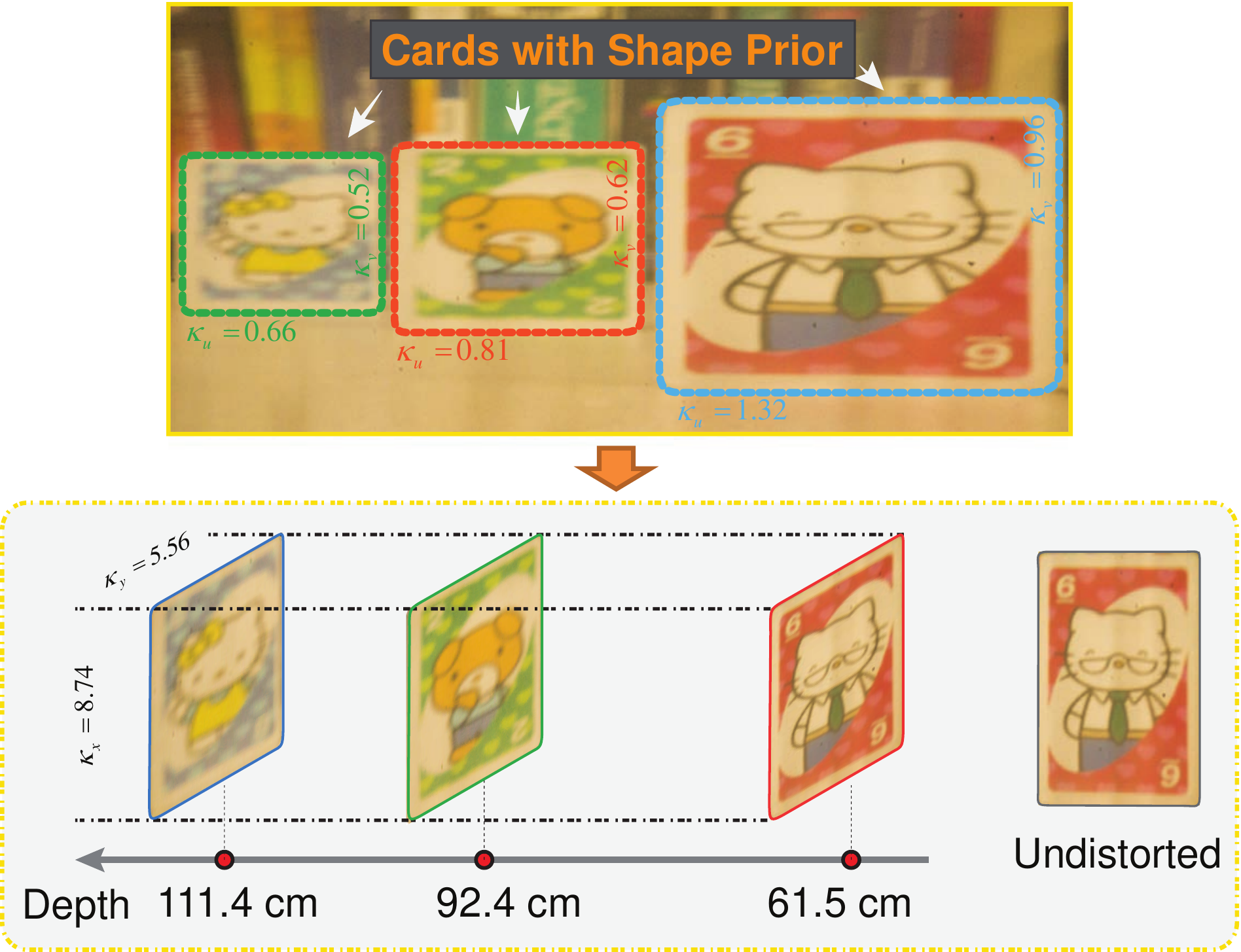}
\centering \caption{Depth-from-DDAR: Top shows a scene that contains multiple cards of an identical but unknown size. Bottom shows their recovered depths and original size using our scheme from this single image.}
\label{fig:CardDemo}
\end{figure}

Our analysis reveals that if we know $r_o$ in prior, i.e., the base aspect of the object, we can directly infer the object's depth from its aspect ratio in the XSlit camera. A typical example is using an Parallel-Orthogonal XSlit camera (PO-XSlit) to capture an up-right rectangle. In a PO-XSlit camera, the slits are orthogonal and axis aligned. In this case, $r_o$ directly corresponds to the aspect ratio of the rectangle and $r_i$ corresponds to the observed AR of the project rectangle.

The simplest case is to capture a up-right square whose aspect ratio $r_o = 1$. From the AR change, we can directly infer its depth using Eqn. ~\ref{eq:getz}. In practice, we do not know the AR of the object in prior. However, many natural scenes contain (rectangular) objects of identical sizes (e.g., windows of buildings) and we can infer their depth even without knowing their ground truth AR.

\paragraph{Shape Prior} Specifically, consider $K$ rectangles of an identical but unknown sizes and hence ARs. Assume they lie at different depths $z^j$. According to Eqn. ~\ref{eq:pinproj}, we have two equations for each rectangle:

\begin{equation}
\label{eq:smshapeprior}
\begin{split}
\kappa_u^j z^j + z_2 \kappa_x = z_2 \kappa_u^j \\
\kappa_v^j z^j + z_1 \kappa_y = z_1 \kappa_v^j \\
\end{split}
\end{equation}

Where $j = 1..K$, $z^j$, $\kappa_x$ and $\kappa_y$ are unknowns. And $\kappa_u$ and $\kappa_v$ are computed from the image. For $K$ identical rectangles, we have $K+2$ unknowns and $2K$ equations. The problem can be solved using SVD when two or more identical rectangles are present. Fig. ~\ref{fig:CardDemo} shows several examples using our technique recovering depth of multiple cards of an identical size. The depth along with the exact scale can be extracted from a single XSlit image under the shape prior.

\paragraph{Depth Prior} If the objects are of identical aspect ratios but of different sizes, still exhibit ambiguity. Then according to Eqn. ~\ref{eq:ardef}, there are $K$ equations and $K + 1$ unknowns (assume $K$ objects). One useful prior that can be imposed here is the distribution of depth of objects. In real scenes, objects are likely to br evenly distributed. For example, if we assume that these rectangles are with equal distance along the $z$ direction.

In this scenario/case, we obtain the AR equation for each object:
\begin{equation}
\label{eq:eqdstprior}
z^jr_o-r_i^j\frac{z_1}{z_2}z^j-z_1r_o = -z_1r_i^j, \hspace{8pt} j=1..K
\end{equation}

Furthermore, the equal distance prior gives us the constraint $z^j - z^{j-1} = z^{j+1} - z^j$, for $j = 2...(K-1)$. For $K$ objects in the scene, we have $2K-2$ equations, and $K+1$ unknowns. The problem is determined if we have 3 rectangles in the scene. And it's over-determined if we have more than 3 objects.

It is very important to note that inferring depth under the same setting is not possible in the perspective camera case. In pinhole image $z_1 = z_2$ and $r_i = r_o$, hence Eqn. ~\ref{eq:smshapeprior} and Eqn. ~\ref{eq:eqdstprior} degenerate. As shown in the introduction, scaling the scene and adjusting the distance from the scene to the pinhole camera accordingly will result in a same projected image as the ground truth scene dose.

\subsection{Line Slope Analysis}
\begin{figure}
\includegraphics[width=0.9\linewidth]{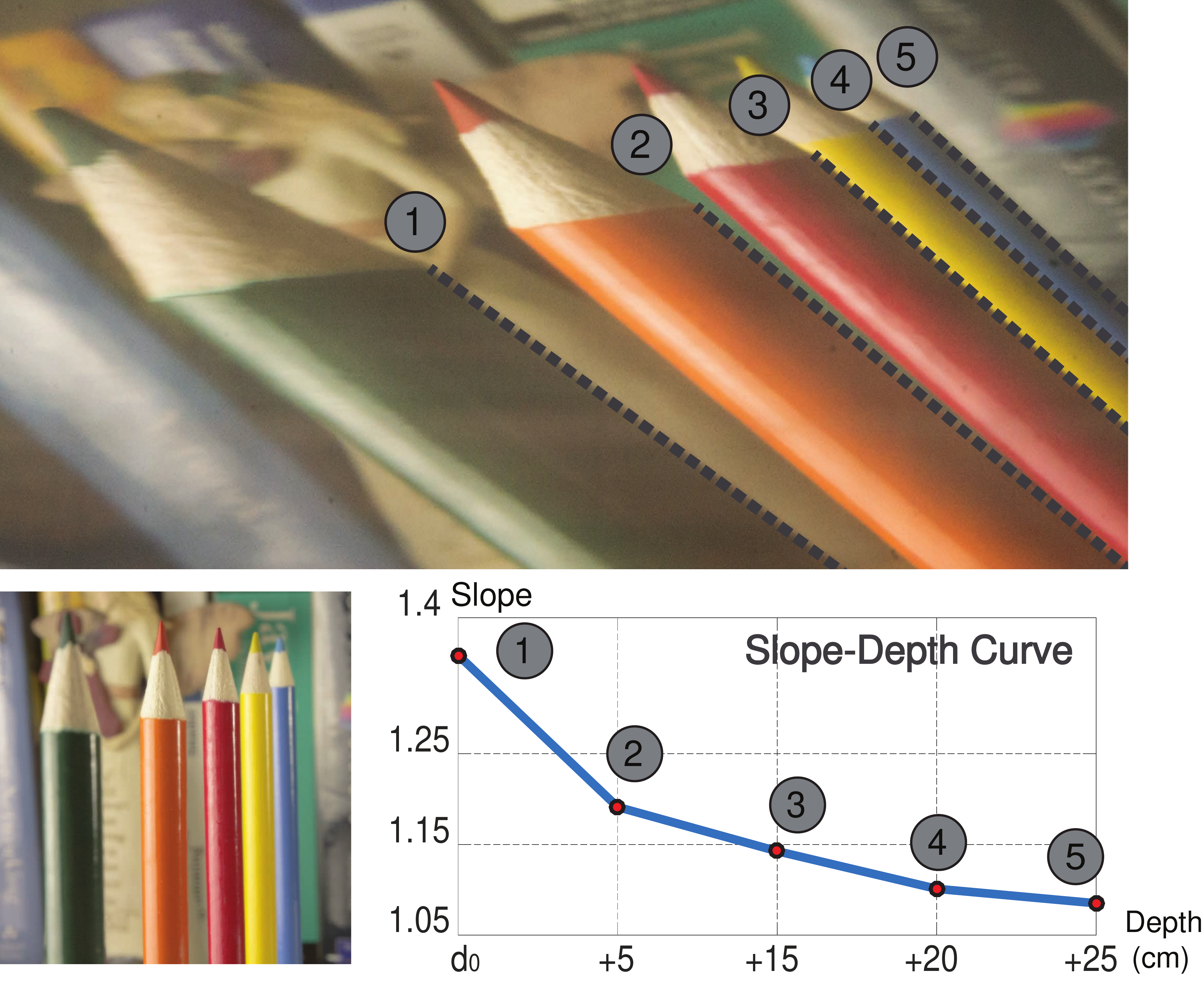}
\centering \caption{Extending DDAR to DDS. Top: parallel 3D lines map to 2D lines of different slopes in an XSlit image. Bottom: the slopes can be used to infer the depths of the lines.}
\label{fig:pencildemo}
\end{figure}

Section ~\ref{sec:depthinfer} reveals that inferring depth from DDAR is that we need to obtain some prior knowledge of either the base AR $r_o$ or the depth distribution of multiple identities. Further, the rectangular shape needs to be in the up-right position to align with the two slits. In this section, we extend the AR analysis to study the slope of lines and we show that this analysis leads to a more effective depth inference scheme.

We treat a line frontal parallel to the XSlit camera as the diagonal of a parallelogram (rectangle in PO-XSlit case), whose sides are along the two slits directions. Given a line with slope $s$ and a point $p_1[x_1, y_1, z]$  on it, then we have $p_2[x_1 + 1, y_1 + s, z]$ of is on the line. We can map it to a line with slope $s^\prime$ on XSlit image, which $p_1$ and $p_2$ map to points $p_1^\prime(u_1,v_1)$ and $p_2^\prime(u_1 + c,v_1 + cs^\prime)$ respectively. According to definition of $r_o$, we can decompose the segment $p_1$-$p_2$ onto two slits direction and take the ratio of the two component to get $r_o$:

\begin{equation}
\label{eq:lro}
r_o = \frac{\sin{\theta_2} - s \sin{\theta_1}}{s\cos{\theta_1} - \cos{\theta_2}}
\end{equation}

$r_i$ is computed using Eqn.~\ref{eq:lro} too, only substitute $s$ with $s^\prime$. Reuse Eqn.~\ref{eq:getz}, we can get the depth.

Eqn.~\ref{eq:lro} and ~\ref{eq:getz} reveals that we can directly infer the depth of the line from its slope. Similar to the aspect ratio case, such inference cannot be conducted in the pinhole camera since the frontal parallel line slope is invariant to depth.

The analysis above applies only to lines parallel to XSlit camera. For lines unparallel to the camera, previous studies have shown that they map to curves, or more precisely hyperbolas \cite{ye2013manhattan}. However, our analysis can still be applied by computing the tangent direction on the hyperbolas, where each tangent direction can be mapped to a unique depth. This can be viewed as approximating a line as piecewise segments frontal-parallel to the camera where each segment's depth can be computed from its projected slope. The complete derivation is included in the supplementary materials.

\subsection{Scene Reconstruction}
\label{sec:mrf}
Based on our theories, we present a new framework on single-image Manhattan scene reconstruction using the XSlit camera. The main idea here is to integrate depth cues from DDAR (for up-right rectangle objects) and from line slopes (for other lines and rectangles) under a unified depth inference framework. Further, the initial depth estimation scheme can only infer depths on pixels lying on the boundaries of the objects, it is important to propagate the estimation to all pixels in order to obtain the complete depth map.

Our approach is to first infer the depth for the lines or repeat objects from DDAR. Next we cluster pixels into small homogenous patches or superpixels \cite{felzenszwalb2004efficient}. The use of superpixels not only reduce the computational cost and but also preserves consistency across the regions, i.e the pixels in a homogeneous region such as walls of a building tend to have a similar depth. Finally, we model optimal depth estimation/propagtion as a Markov Random Field (MRF). The initial depth value $V_i$ for superpixel $S_i$ is computed by blending the depths inferred from DDAR according to their geodesic distance to $S_i$. And then we the smooth out $V$ based on distance variations and color consistency. This procedure can be modeled as a Markov Random Field (MRF), where the data term: $E_d(S_i) =U_i - V_i$. And the smoothness term is: $E_s(S_i, S_j) = w_{ij}(U_i - U_j)$, $w_{ij}$ is the weight account for distance variations and color consistency. Finally we estimate the depth map $U$ by optimizing the energy function: $E(U) =  \sum \limits_{S_i} E_d(Si) + \lambda \sum \limits_{S_i, S_j \in N} E_s(S_i, S_j)$, $N$ represents the superpixel neighborhood. The problem can be solved using the graph-cut algorithm \cite{boykov2004experimenta}.

%XXX: just add one equation.

%discuss how you solve the MRF. graph-cut or belief propagation. how to discretize.

%Assume we have $N$ detected depths from line or object $l$ depth is at depth $d_l$. For each superpixel $S_i$, we have the energy function:
%
%\begin{equation}
%\label{eq:kuv}
%E(S_i) = E_d(S_i) + \lambda E_s(S_i)
%\end{equation}
%
%Where
%\begin{equation}
%\begin{split}
%\label{eq:kuv}
%E_d(S_i) = xxxx \\
%E_sS_i) = xxxx \\
%\end{split}
%\end{equation}
%
%$\lambda$ is used to adjust the degree of smoothing.

\section{Experiments}
We experiment our approach on both synthetic and real scenes. For synthetic scenes, we render images using 3ds Max. For real scenes, we acquire images using the XSlit lens as well as synthesize XSlit panoramas from video sequences.

\paragraph{Synthetic Results.}

\begin{figure}
\includegraphics[width=1.0\linewidth]{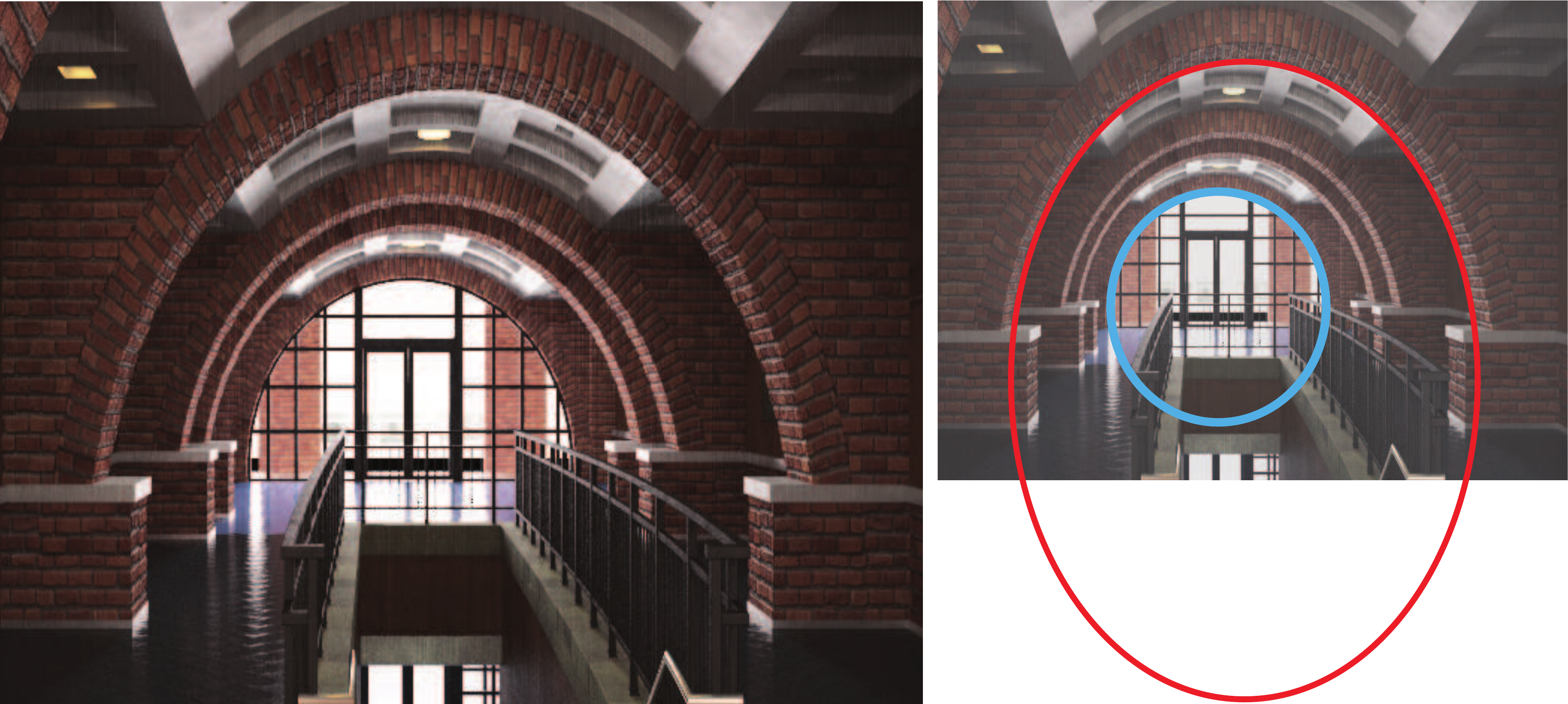}
\centering \caption{An XSlit image of the arch scene that contains 3D concentric circles (left). Their images correspond to ellipses of different aspect ratios (right). }
\label{fig:circlescene}
\end{figure}

We first render an XSlit images of a scene containing repeated shapes (Fig.~\ref{fig:circlescene}). The architecture consists of concentric arches of depths ranging from 900cm to 2300cm. We assume that the actual aspect ratio of the arches is 1, i.e., a circle. We position a PO-XSlit camera with $z_1 = -3.2$cm and $z_2 = -346.7$cm frontal parallel to the arches and the images of the arches are ellipses of different aspect ratios. Notice that in the pinhole case, they will be map to circles. We first detect ellipses using Hough transform and then measure their aspect ratios using the major and minor axes. Finally, we use the ratios to recover their depths using Eqn.~\ref{eq:getz}. Our recovered depths for the near and far arches are 906.6cm and 2281.0cm, i.e., the errors are less than 2\%.

\begin{figure*}
\includegraphics[width=1.0\linewidth]{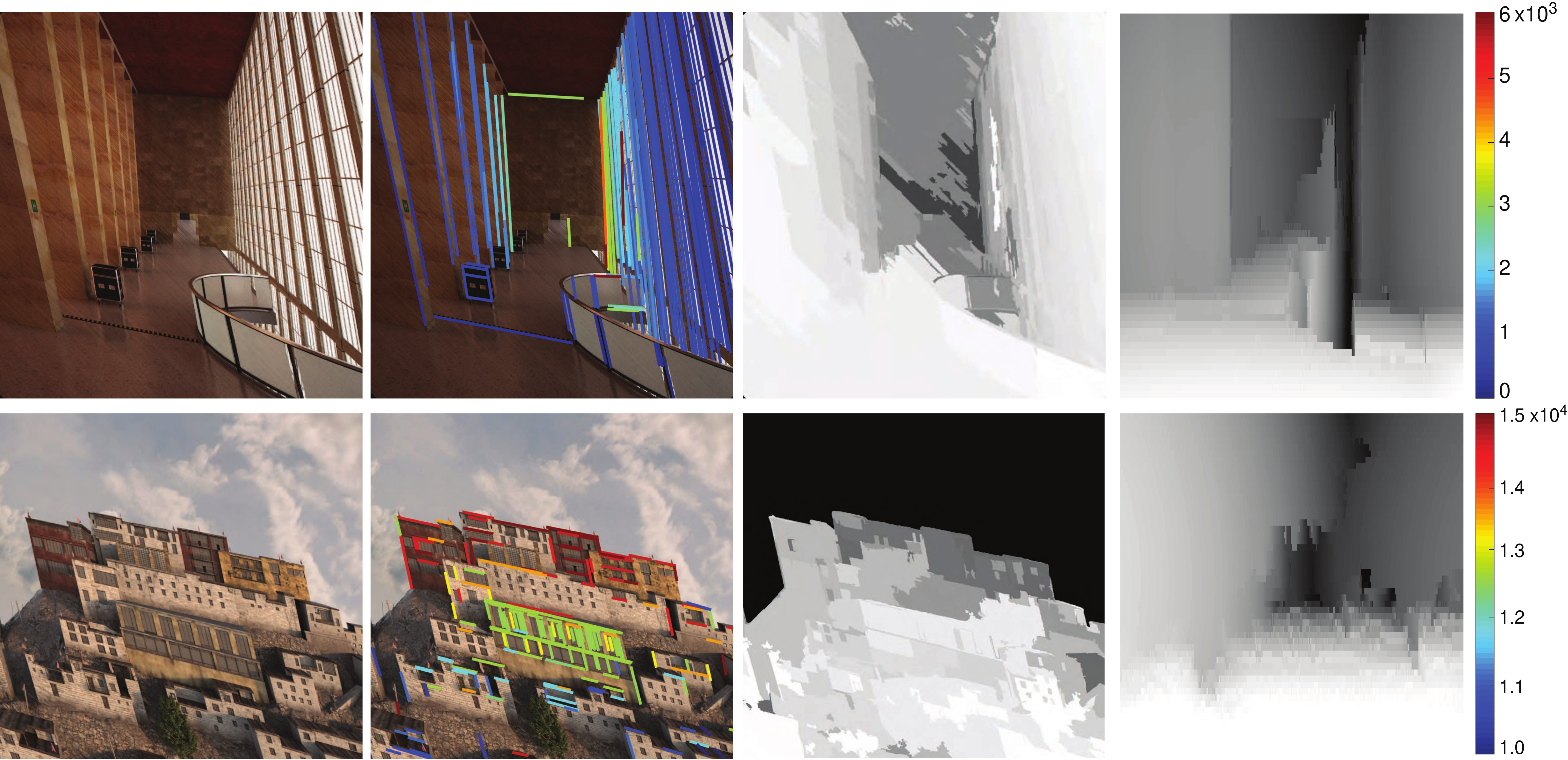}
\centering \caption{From left to right: An XSlit image of a scene containing parallel 3D lines, the detected lines and their estimated depth using DDS, the depth map recovered using our scheme, and the one recovered using Make3D \cite{saxena2007learning} by using a single perspective image. }
\label{fig:maxscene}
\end{figure*}

Next we render two XSlit panoramas, one for the corridor and the second for the facade. Both scenes exhibit strong linear structures with many horizontal and vertical lines. Our analysis shows that for lines to exhibit DDS, they should not align with either slit. Therefore, we rotate the POXSlit, i.e., $\theta_1 = 45^\circ$ and $\theta_2 = 135^\circ$. For the corridor scene, the XSlit camera has a setting of $z_1 = -3.6$cm, $z_2 = -717.9$cm and for the facade scene, $z_1 = -3.1$cm, $z_2 = 4895.9$cm. We first use the LSD scheme\cite{von2012lsd} to extract 2D lines from the XSlit images and cluster them into groups of horizontal and vertical (in 3D) lines. This is done by thresholding their aspect ratios  Eqn.~\ref{eq:ratiorange}. For lines in each group, we compute their depths using Eqn.~\ref{eq:lro} and ~\ref{eq:getz}. This results in a sparse depth map. To recover the full depth map, we apply the MRF (Sec.~\ref{sec:mrf}) and the final result is shown in Fig.~\ref{fig:maxscene}. Our technique is able to recover different depth layers while preserving linear structures. For comparison, we render a single perspective image and apply the learning-based scheme Make3D \cite{saxena2007learning}. Make3D can detect several coarse layers but cannot detect fine details as ours since these linear structures appear identical in slope in a perspective image but exhibit different slopes in an XSlit image.

%------------------------------------------------------------------------
\paragraph{Real Results.}

We explore several approaches to acquire XSlit images of a real scene: by a real XSlit lens and through panorama synthesis. For the former, we use an XSlit lens \cite{ye2013manhattan}. The design resembles the original anamorphoser proposed by Ducos du Hauron that replaces the pinhole in the camera with a pair of narrow, perpendicularly crossed slits. Similar to the way of using a spherical thin lens to increase light throughput in a pinhole camera, the XSlit lens relay perpendicular cylindrical lenses, one for each slit. In our experiments, we use two cylindrical lenses with focal lengths 2.5cm (closer to the sensor) and 7.5cm (farther away from the sensor) respectively. The distance between the two slits is adjustable between 5cm and 12cm and the slit apertures have a width of 1mm.

\begin{figure}
\includegraphics[width=0.9\linewidth]{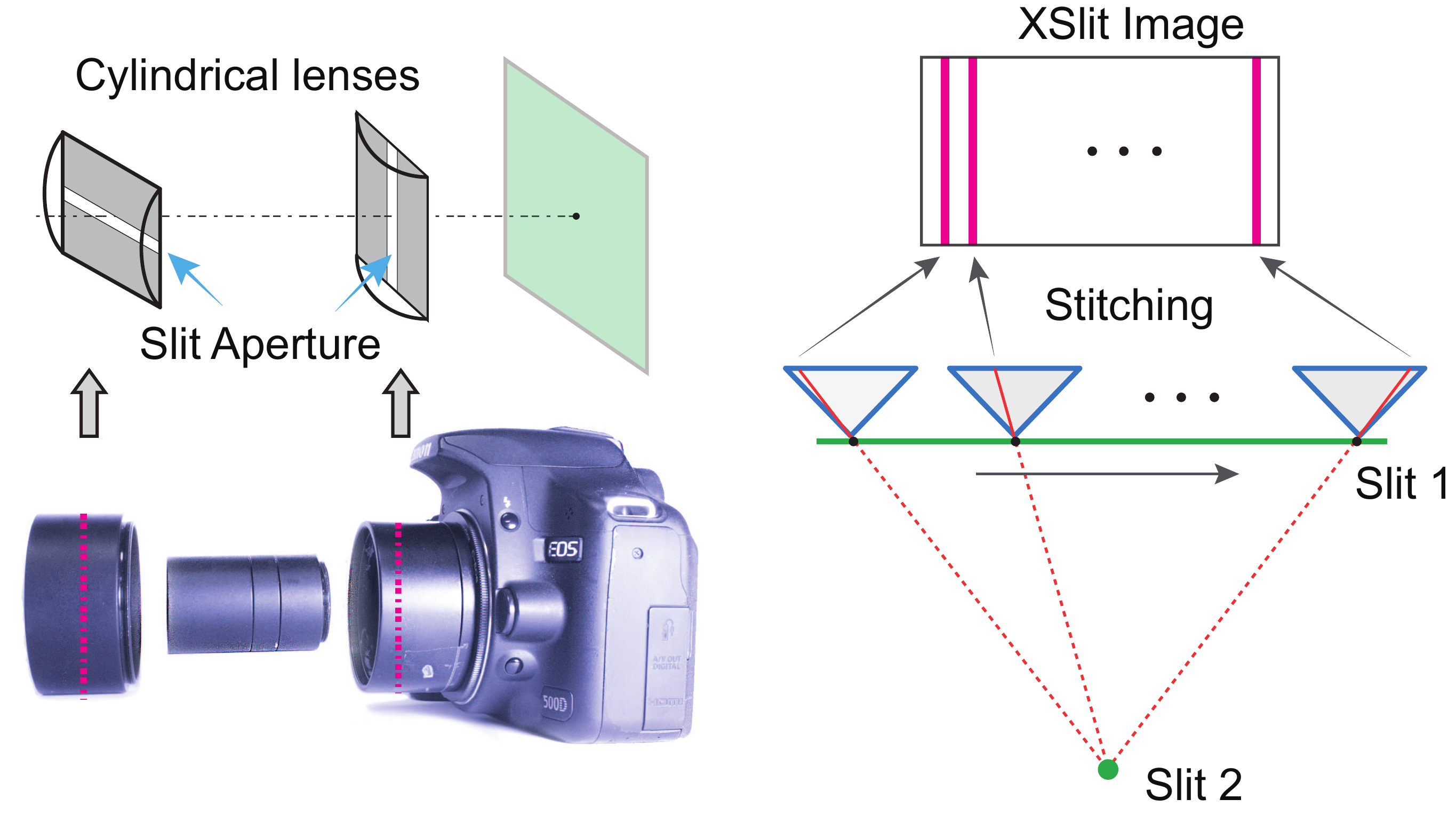}
\centering \caption{ XSlit images can be captured by a real XSlit lens (left) or by stitching linearly varying columns from a 3D light field (right).}
\label{fig:genxslits}
\end{figure}

\begin{figure}
\includegraphics[width=0.89\linewidth]{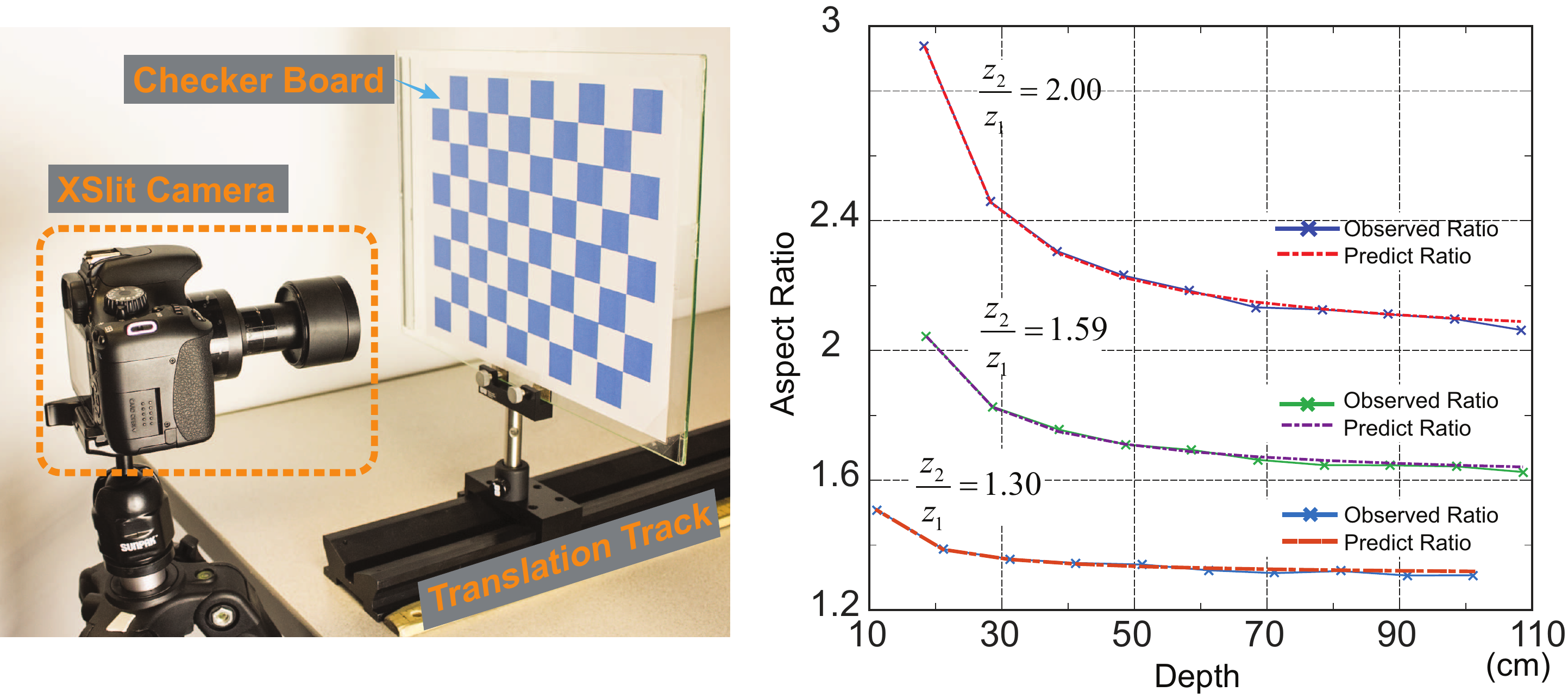}
\centering \caption{Experimental validations of our analysis. We place checker board in front of the XSlit camera and move it away(Left). The comparisons of measured AR and predict AR with different silts configurations(Right).}
\label{fig:RealExTheo}
\end{figure}

We first capture a checkerboard at known depths and compare the measured AR and our predicted AR using Eqn.~\ref{eq:getz}. We test three different slit configurations, $z_2/z_1 = 1.3$, $z_2/z_1 = 1.59$ and $z_2/z_1 = 2.0$. Fig.~\ref{fig:RealExTheo} shows that the predicted AR curve fits well with the ground truth. In particular, as an object gets farther away from the sensor, its AR also changes slower. Further, the larger the baseline $z_2/z_1$ is, the larger the aspect ratio variations across the same depth range, as predicted by our theory.

%Next, we experiment on the two depth recovery schemes, one based on depth prior and the second on verifying the line slope analysis. Fig. XXX shows a nesting doll scene composed of 3 Santa dolls at different depths. The dolls are with same aspect ratio but different size. The distance between the three dolls is 15cm. Fig. XXX shows the depth estimation using the depth prior. The ground truth aspect ratio of the doll is 2.39. The predict aspect ratio is 2.45. The computed separation distance is 16.02cm. Our predict is consistent with the ground truth. %Discuss how accuracy decreases with respect to depth, which is not surprising. Fig. XXX shows that our reconstruction result if we assume that the AR is unknown but the cards are placed at evenly separated depth. Compare both the card's dimension and depth recovered from our algorithm.

\begin{figure}
\includegraphics[width=1.0\linewidth]{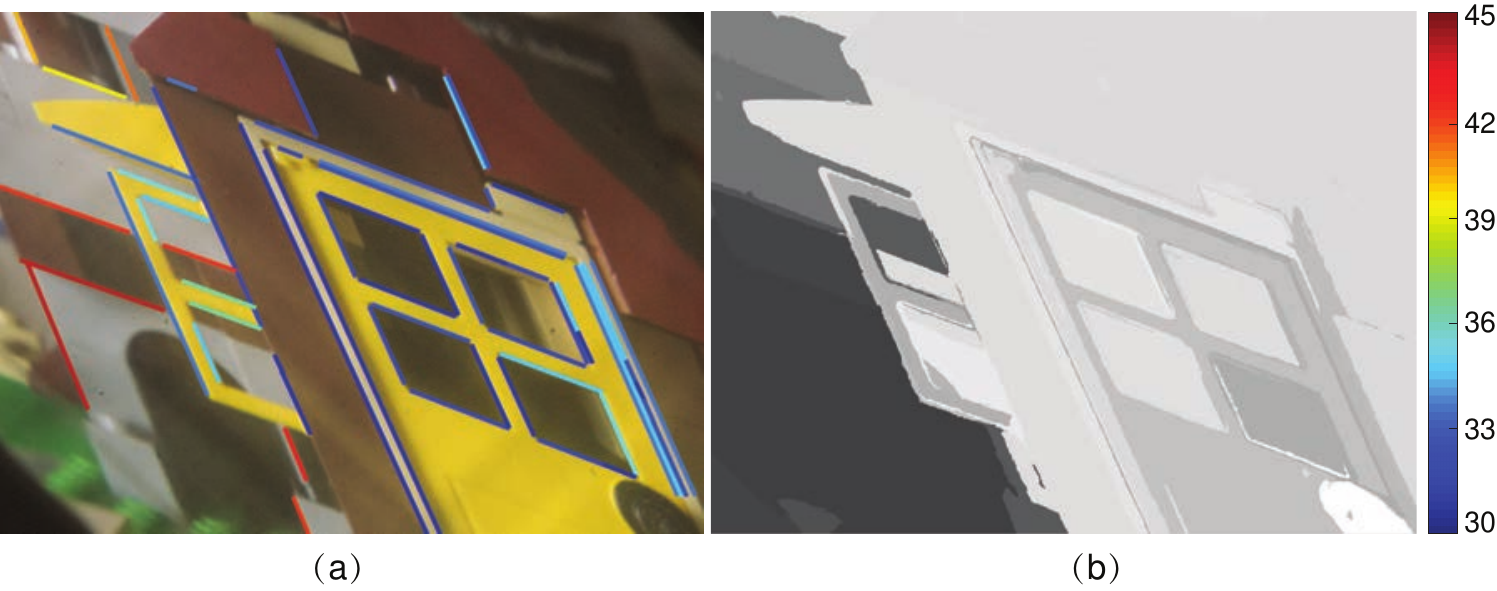}
\centering \caption{Real result on a Lego$^{\circledR}$ house scene. (a) an XSlit image of the scene captured by the XSlit camera. Detected lines are highlighted in the image. (b) the recovered depth map using our slope and aspect ratio based scheme. }
\label{fig:legoex}
\end{figure}

Next, we verify our DDS analysis using images captured the XSlit camera. In Fig.~\ref{fig:legoex}, we position a Lego$^{\circledR} $ house model in front of the XSlit camera ($z_1 = 6.12$cm and $z_2 = 11.81$cm). We rotate the XSlit camera by 45 degrees so that the 3D lines on the house will not align with either slit. Fig.~\ref{fig:legoex}(a) shows the acquired image. Next, we conduct line fitting and slope estimation similar to the synthetic case for estimating the depths of the detected lines. Fig.~\ref{fig:legoex}(a) highlights the detected lines and their depths (using color) and Fig.~\ref{fig:legoex}(b) shows the complete depth map using the MRF solution. The results shows that major depth layers are effectively recovered. The error on the top-right corner is caused by the lacking of line structures.

A major limitation using the XSlit camera is its small baseline (between the two slits). Our analysis shows that the maximum recoverable depth range depends on this baseline. Further, since images captured by the XSlit camera exhibits noise and strong defocus blurs, the actual recoverable depth range is even smaller. For example, our analysis shows that with baseline $z_2/z_1 = 2$, two cards are placed at $30m$ and $35m$ will have undistinguishable ARs. Their ratio difference reach the lower bound that determined by pixel size. For outdoor scenes, we resort to XSlit panorama synthesis.

\begin{figure}
\includegraphics[width=1.0\linewidth]{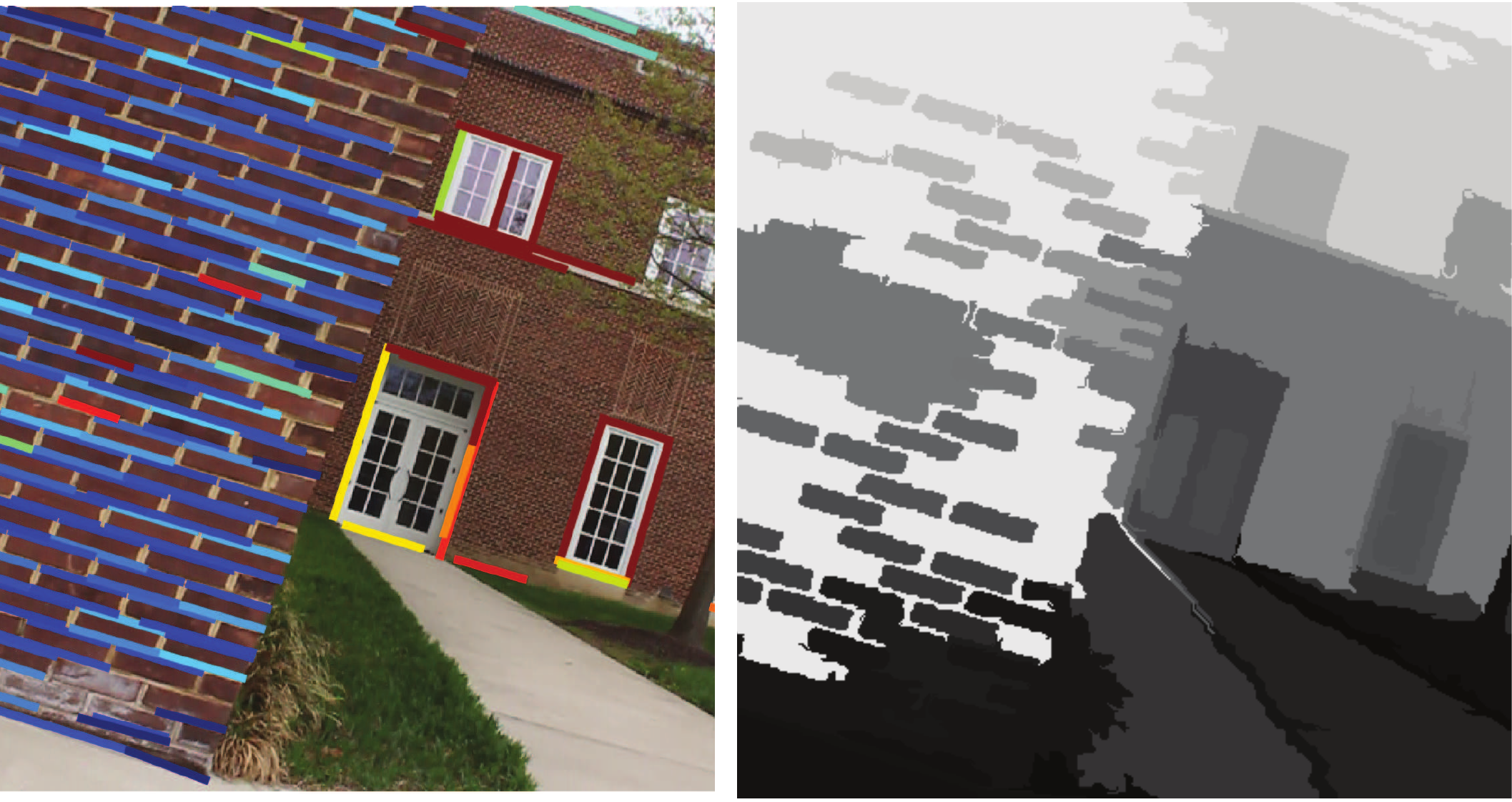}
\centering \caption{The XSlit image of an outdoor scene. Left: An XSlit panorama and the detected lines. Right: The recovered depth map.}
\label{fig:outdoor}
\end{figure}

To produce XSlit panoramas, Zomet et al. \cite{Zomet09XSlit} capture a sequence of images by translating a pinhole camera along a linear trajectory at a constant velocity. In a similar vein, Seitz and Adams et al. acquire the image sequence by mounting the camera on a car facing towards the street. Additional registration steps \cite{AgarwalaACSS06} can be applied to rectify the input images. Next, linearly varying columns across the images are selected and stitched together. Fig.~\ref{fig:genxslits} shows the procedure of generating a XSlit image using a regular camera.

\begin{figure*}
\includegraphics[width=0.75\linewidth]{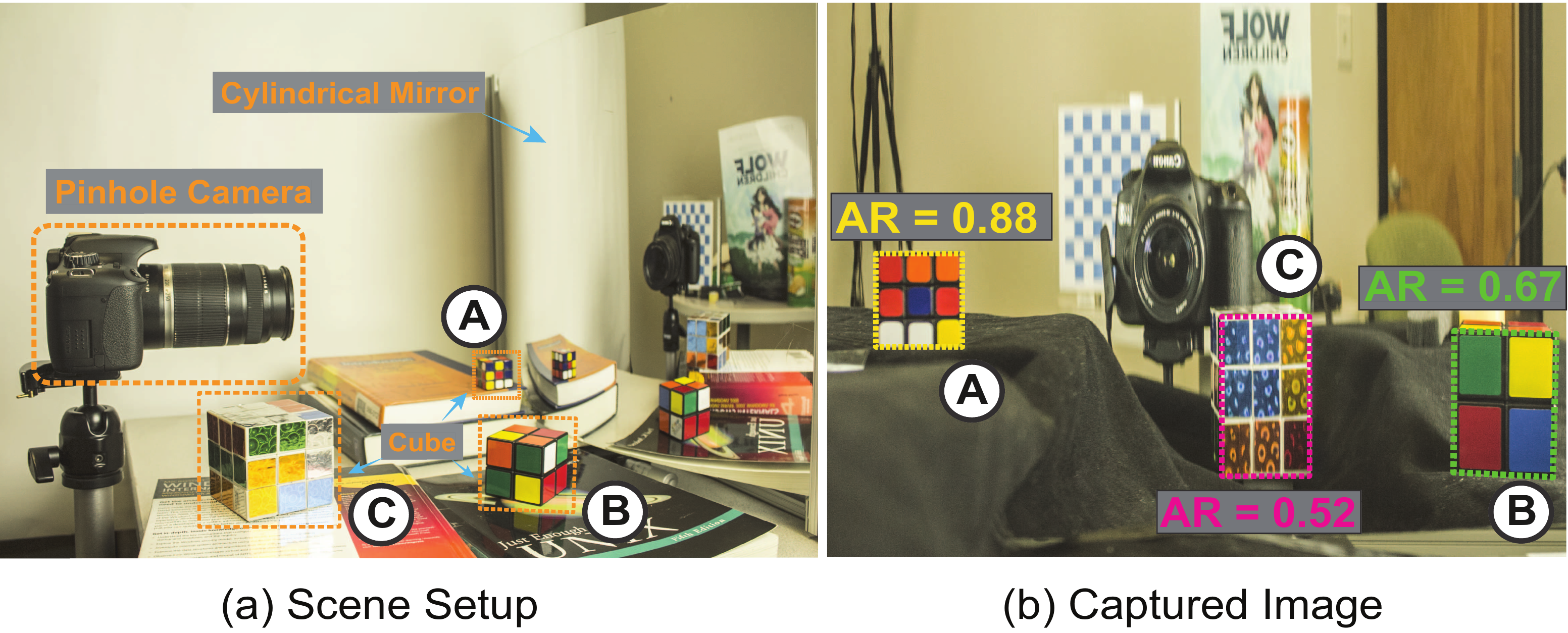}
\centering \caption{Results on catadioptric mirrors. Left: we capture the scene using a cylindrical catadioptric mirror. Right: the aspect ratios of cubes change with respect to their depthes.}
\label{fig:cyldemo}
\end{figure*}

Fig.~\ref{fig:outdoor} shows the XSlit panorama synthesized from an image sequence captured by a moving camera. We linearly increase the column index in terms of frame number and stitch these columns to form an XSlit image. The moving path of the camera is 55cm long. And the camera is tilt with 20$^\circ$ angle. The resulting two slits are at -1.8cm and 41cm respectively.

Recent ray geometry studies \cite{ding2009recovering} show that reflections of certain types of catadioptric mirror can be approximated as an XSlit image. In Fig. ~\ref{fig:cyldemo}, we position a perspective camera facing towards a cylindrical mirror and Fig. ~\ref{fig:cyldemo}(b) shows that DDAR can both be observed on the acquired image. In particular, we put multiple cubes of an identical size at different depths and their aspect ratios change dramatically. This is because two virtual slits of the catadioptric mirror are separated far away where DDAR is more significant than the XSlit camera case. .

%-------------------------------
\section{Conclusion and Further Work}
We have comprehensively studied the aspect ratio (AR) distortion in XSlit cameras and exploited its unique depth-dependent property for 3D inference. Our studies have shown that unlike perspective camera that preserves AR under depth variations, AR changes monotonically with respect to depth in an XSlit camera, i.e., 3D objects of an identical size will exhibit significantly different AR under different depths. This has led to new depth-from-AR schemes using a single XSlit image even if the original AR of an object is unknown. We have further shown that similar to AR variations, the slope of projected 3D lines will also vary with respect to depth, and we have developed theories to characterize such variations based on AR analysis. Finally, AR and line slope analysis can be integrated for 3D reconstruction and we have experimented on real XSlit images captured by an XSlit camera, synthesized from panorama stitching, and captured using a catadioptric mirror to validate our framework.

There are a number of future directions we plan to explore. Our cylindrical lens based XSlit has a small baseline (i.e., the distance between the two slits) and therefore can only acquire AR changes within a short range. Constructing a large baseline XSlit camera will be costly as it is difficult to fabricate large form cylindrical lens. A more feasible solution would be adopt a cylindrical catadioptric mirror where the reflection image can be approximated as an XSlit image. In the future, we will explore effective schemes for correcting both geometric distortion and blurs due to imperfect mirror geometry. We will also investigate integrating our AR based solution into prior based frameworks to enhance reconstruction quality. For example, a hybrid XSlit-perspective camera pair can be constructed. Finally, since AR distortions commonly exhibit in synthesized panoramas as shown in the paper, we plan to study effective image-based distortion correction techniques to produce perspectively sound panoramas analogous to \cite{AgarwalaACSS06}.

{\small
\bibliographystyle{ieee}
\bibliography{arbib}
}

\end{document}